\documentclass[10pt,twocolumn,letterpaper]{article}

\usepackage[pagenumbers]{cvpr} % To force page numbers, e.g. for an arXiv version
\usepackage{graphicx}
\usepackage{amssymb}
\usepackage{siunitx}
\usepackage[ruled,vlined]{algorithm2e}
\usepackage{wrapfig}

\definecolor{cvprblue}{rgb}{0.21,0.49,0.74}
\usepackage[pagebackref,breaklinks,colorlinks,allcolors=cvprblue]{hyperref}

\def\paperID{} % *** Enter the Paper ID here
\def\confName{}
\def\confYear{}

\title{Domain Adaptation for Different Sensor Configurations in 3D Object Detection}

\author{Satoshi Tanaka \thanks{Equal contribution} \quad , Kok Seang Tan \footnotemark[1] \quad, Isamu Yamashita \\
TIER IV, Inc\\
{\tt\small satoshi.tanaka@tier4.jp}
}

\begin{document}
\maketitle

\begin{abstract}
Recent advances in autonomous driving have underscored the importance of accurate 3D object detection, with LiDAR playing a central role due to its robustness under diverse visibility conditions.
However, different vehicle platforms often deploy distinct sensor configurations, causing performance degradation when models trained on one configuration are applied to another because of shifts in the point cloud distribution.
Prior work on multi-dataset training and domain adaptation for 3D object detection has largely addressed environmental domain gaps and density variation within a single LiDAR; in contrast, the domain gap for different sensor configurations remains largely unexplored.
In this work, we address domain adaptation across different sensor configurations in 3D object detection.
We propose two techniques: Downstream Fine-tuning (dataset-specific fine-tuning after multi-dataset training) and Partial Layer Fine-tuning (updating only a subset of layers to improve cross-configuration generalization).
Using paired datasets collected in the same geographic region with multiple sensor configurations, we show that joint training with Downstream Fine-tuning and Partial Layer Fine-tuning consistently outperforms naive joint training for each configuration.
Our findings provide a practical and scalable solution for adapting 3D object detection models to the diverse vehicle platforms.

\end{abstract}

\section{Introduction}
\label{sec:intro}

\begin{figure*}[t]
   \begin{center}
      \includegraphics[width=0.98\linewidth]{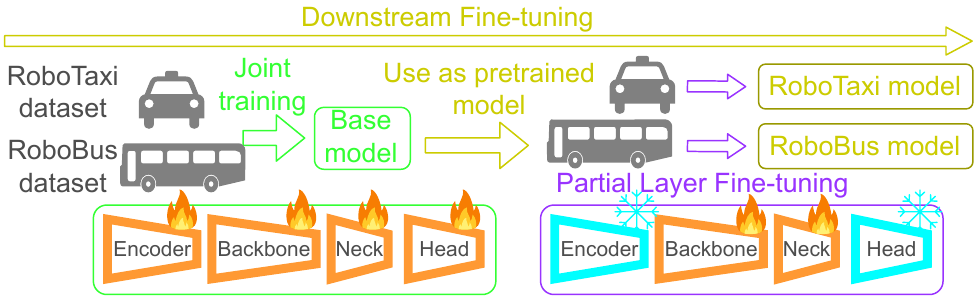}
   \end{center}
   \caption{
      Overview of the proposed training strategy combining Downstream Fine-tuning and Partial Layer Fine-tuning, as our concept of domain adaptation for different sensor configurations.
      Downstream Fine-tuning improves performance by first jointly training on all sensor configurations to learn generalizable features, then fine-tuning each configuration to specialize to its characteristics.
      Partial Layer Fine-tuning improves performance by selectively updating layers most sensitive to sensor-configuration shifts — training the backbone and neck while freezing the encoder and head — to adapt spatial features.
   }
   \label{concept}
\end{figure*}

The rapid development of autonomous driving has emphasized the critical role of 3D object detection in understanding complex environments \cite{Chen20203DPC, mao20233dobjectdetectionautonomous}.
Among various sensing modalities, LiDAR (Light Detection and Ranging) has emerged as a primary sensor for autonomous vehicles due to its ability to provide high-resolution distance information and maintain robustness in challenging visibility conditions.
Unlike conventional 2D image-based detection, LiDAR-based 3D object detection enables precise estimation of object positions, shapes, and categories in 3D space, offering richer spatial awareness.
Importantly, autonomous vehicles are not limited to passenger cars \cite{waymo, ponyai} — they also include diverse platforms such as buses \cite{zoox, weride}, trucks \cite{tusimple, plus}, and last-mile delivery robots \cite{meituan, nuro}.
Each platform often employs a distinct sensor configuration, which significantly affects the distribution, density, and coverage of the resulting point clouds.
As a result, a 3D object detection model optimized for a particular sensor setup often suffers from severe performance degradation when applied to data collected with a different configuration.
Given the real-world deployment of autonomous systems with heterogeneous sensor layouts, it is imperative to enhance the generalization ability of 3D detectors across varying sensor configurations.

To address domain shifts in other contexts, such as geographic or environmental changes, research in domain adaptation for 3D object detection has gained increasing attention \cite{Triess2021ASO}.
However, prior work has largely overlooked the sensor-configuration-induced domain gap.
In practice, the placement and number of LiDAR sensors differ substantially: for example, nuScenes features a single top-mounted LiDAR, while commercial bus platforms often adopt corner-mounted configurations to mitigate self-occlusion.
Even within the same geographic region, these structural differences result in distinct point cloud characteristics, leading to sensor-specific domain gaps.

An alternative approach to improve generalization is multi-dataset training, in which multiple labeled datasets from different domains are jointly used for training.
While this strategy can produce robust models, it often leads to trade-offs in per-dataset performance due to overgeneralization — a phenomenon reminiscent of the No Free Lunch theorem.
Notably, in many robotics applications, especially those deployed in fixed environments, it is acceptable to deploy per-configuration models rather than enforce full generalization across all sensor types.
Despite this practical relevance, research specifically aimed at obtaining high-performing models for each sensor configuration remains limited.

Motivated by this domain gap, we investigate domain adaptation for 3D object detection across different sensor configurations.
An overview of our proposed framework is presented in Figure \ref{concept}.
We introduce two adaptation techniques: Downstream Fine-tuning and Partial Layer Fine-tuning.
Downstream Fine-tuning adopts a two-stage training strategy, where a model is initially trained on multiple datasets and subsequently fine-tuned separately on each dataset.
Partial Layer Fine-tuning updates only a subset of layers, enabling efficient and configuration-specific adaptation.
To support our experiments, we construct a dataset comprising sensor configurations for both RoboTaxi and RoboBus, collected within the same geographic region.
We evaluate the proposed methods under both sensor configuration settings.
Specifically, we make three key contributions:
{
\begin{itemize}
\setlength{\itemsep}{-0.3mm}
\item We construct a dataset featuring different sensor configurations that share a common annotation specification, collected within the same geographic domain, to analyze the domain gap introduced by sensor variation.
\item We propose a supervised training pipeline combining Downstream Fine-tuning and Partial Layer Fine-tuning to adapt 3D detectors for each sensor configuration individually.
\item We construct a new multi-configuration 3D detection benchmark with different sensor setups, and provide a thorough quantitative evaluation.
\end{itemize}
}

\section{Background}
\label{sec:background}

\begin{table}[t]
    \centering
    \small
    \begin{tabular}{|c|c|c|c|c|}
    \hline
    & KITTI & ONCE & nuScenes & Waymo \\
    \hline
    \shortstack{Single dataset} & 47.4 & 60.2 & 18.4 & 40.2 \\
    \hline
    \shortstack{MDT3D \cite{MDT3D}} & 41.8 & 38.2 & 11.0 & 6.4 \\
    \hline
    \end{tabular}
    \caption{
        Multi-dataset training underperforms compared to models trained on each single dataset individually.
        }
\label{tab:no_free}
\end{table}

\begin{table}[t]
    \centering
    \scriptsize
    \setlength{\tabcolsep}{2pt}
    \begin{tabular}{|l|c|c|c|c|}
    \hline
    & UDA, SSDA & TTA & MDT & Ours \\
    \hline
    Annotation configuration & Different & Different & Different & Same \\
    Sensor configuration & Similar & Similar & Similar & Different \\
    Source data & $\{x^s, y^s\}$ & None & $\{x^s, y^s\}$ & $\{x^s, y^s\}$ \\
    Target data & $\{x^t\}$ (+$\{y^t\}$) & $x^t$ & $\{x^t, y^t\}$ & $\{x^t, y^t\}$ \\
    Improve for target data? & Yes & Yes & Yes & Yes \\
    Improve for source data? & No & No & Yes & Yes \\
    \hline
    \end{tabular}
    \caption{Method comparison between unsupervised domain adaptation (UDA), Semi-supervised domain adaptation (SSDA), Test-time adaptation
(TTA), and multi-dataset training (MDT).}
\label{compare_work}
\end{table}

\textbf{LiDAR-based 3D object detection.}
For a long time, researchers have developed 3D object detection models using LiDAR point clouds \cite{Qi2016PointNetDL, Qi2017, Yang2018PIXORR3, Shi2018PointRCNN3O,Shi2019PVRCNNPF, Yang20203DSSDP3}.
VoxelNet \cite{Zhou2017VoxelNetEL} introduces a voxel-based end-to-end pipeline for 3D object detection from point clouds.
SECOND \cite{Yan2018SECONDSE} improves voxel-based 3D object detection by introducing sparse 3D convolutions, enabling more efficient and scalable processing of large point clouds.
PointPillars \cite{Lang2018PointPillarsFE} encodes point clouds into a pseudo-image using pillar-based representations.
CenterPoint \cite{yin2021center} formulates 3D object detection as a center point prediction task, enabling accurate and efficient localization of objects in point clouds.
TransFusion \cite{Bai2022TransFusionRL} introduces an attention-based framework for the head of 3D object detection, improving performance by multi-modal features.

\textbf{Datasets for 3D domain adaptation.}
There are several datasets commonly used in research on domain adaptation for 3D object detection.
The KITTI dataset \cite{Geiger2012CVPR} was collected in Germany using a 64-beam LiDAR sensor, and contains annotations for 3 classes with a front-view-only.
The nuScenes dataset \cite{nuscenes} was collected in the USA and Singapore using a 32-beam LiDAR sensor, and provides 360-degree annotations for 10 classes.
The Waymo Open Dataset \cite{Waymo2020} was collected in the USA using a 64-beam LiDAR sensor, and includes 360-degree annotations for 5 classes.
The ONCE dataset \cite{mao2021one} was collected in China using a 40-beam LiDAR sensor, and contains 360-degree annotations for 9 classes.
These datasets were acquired with a primary LiDAR sensor on the top of a passenger car.

\textbf{Domain adaptation for 3D object detection.}

Researchers have tackled this problem as an unsupervised domain adaptation (UDA) task using open datasets \cite{yang2021st3d, yang2021st3d++, Xu_2021_ICCV, xu2023revisiting, li2024domain, MS3Dpp, wang2023ssda3d}.
ST3D \cite{yang2021st3d} performs unsupervised domain adaptation for 3D object detection via progressive self-training with pseudo-labels.
ST3D++ \cite{yang2021st3d++} enhances ST3D by applying denoising strategies to improve pseudo-labels in unsupervised domain adaptation for 3D object detection.
Several studies have focused on the differences in object distributions across domains \cite{Zhang_2021_CVPR, wang2020train, lu2024dali}.
Other studies have also focused on differences in LiDAR sensor characteristics, such as the number of channels in mechanical LiDARs and variations across sensor models \cite{hu2023density, wozniak2024uada3d, peng2023cl3d, deng2024cmd}.
Semi-supervised domain adaptation (SSDA) for 3D object detection has also been explored to address the domain gap using a small amount of annotated target data and unlabeled data \cite{wang2023ssda3d, yuan2023bi3d, TODA}.
Test-time adaptation (TTA) for 3D object detection aims to improve 3D object detection without requiring source data or additional annotations \cite{YuanZGYSQC24}.
It enhances bounding box regression through consistency regularization and adaptive optimization during inference.

The framework of multi-dataset training (MDT) for 3D object detection also aims to achieve greater generalization across diverse domains \cite{MDT3D, Wu_2023_ICCV, wang2024one}.
However, as summarized by the experimental results of MDT3D \cite{MDT3D} using CenterPoint in Table \ref{tab:no_free}, show that although training on all datasets jointly improves generalization, it often leads to lower performance on individual datasets compared to models trained solely on those datasets.
This reflects a well-known issue as No Free Lunch theorem.
When the goal is to maximize performance on a specific dataset, which is often the case in industry applications, such joint training may actually degrade performance.

While prior work has advanced, few studies have addressed performance degradation caused by sensor configuration differences.
Most studies use open datasets with differing annotation schemes, focusing on domain gaps in labeling or object size, and often on the number of beams in a single LiDAR.
To our knowledge, this is the first work to directly target the performance impact of sensor placement differences between single- and multi-LiDAR setups, under a unified annotation format.
We summarize the related work in Table \ref{compare_work}.

\section{Method}
\label{sec:method}

\begin{figure*}[t]
   \begin{minipage}{0.49\linewidth}
   \begin{center}
      \includegraphics[width=0.97\linewidth]{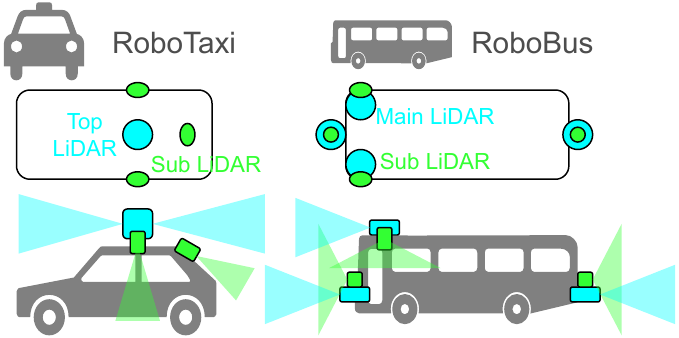}
      \hspace{0.3cm} (a)
   \end{center}
   \end{minipage}
   \begin{minipage}{0.49\linewidth}
   \begin{center}
      \includegraphics[width=0.97\linewidth]{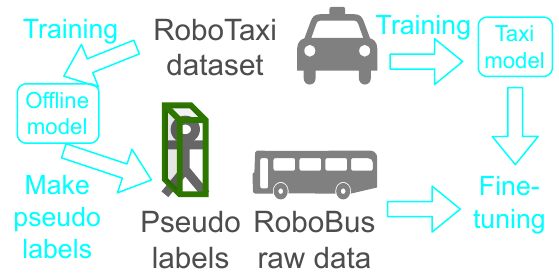}
      \hspace{0.3cm} (b)
   \end{center}
   \end{minipage}
   \caption{
      (a) Multi-configuration dataset with a consistent annotation format across different sensor setups.
      The dataset includes RoboTaxi and RoboBus configurations, reflecting real-world autonomous vehicle platforms with distinct LiDAR layouts, collected across urban and suburban areas in Japan.
      (b) Overview of the unsupervised domain adaptation pipeline to analyze domain gaps from different sensor configurations.
      Pseudo-labels generated by a high-capacity offline model are used to fine-tune the model for a target sensor configuration.
   }
\label{fig:t4dataset}
\end{figure*}

\subsection{Multi-sensor configuration dataset}
\label{sec:dataset}

As illustrated in Figure \ref{fig:t4dataset} (a), we constructed a multi-configuration dataset with a consistent annotation format across different sensor setups.
The dataset includes two representative LiDAR configurations inspired by real-world autonomous vehicle platforms: RoboTaxi and RoboBus.
The RoboTaxi configuration reflects a typical sensor layout for autonomous passenger cars.
It consists of a single top-mounted LiDAR and several surrounding sub LiDARs positioned to minimize blind spots.
All LiDAR point clouds are concatenated and used as the model input.
In our setup, the top LiDAR is a 128-beam sensor, and the sub LiDARs are 16-beam sensors.
This configuration closely resembles existing datasets such as nuScenes.
The RoboBus configuration, in contrast, reflects the sensor layout for autonomous buses.
It includes LiDARs placed at the front, rear, and front corners of the vehicle.
This configuration is necessary for large, box-shaped vehicles like buses, where a single roof-mounted LiDAR would leave substantial blind spots along the lower side and rear areas.
The RoboBus configuration includes multiple 40-beam main LiDARs and multiple 64-beam short-range LiDARs.
All data for the RoboTaxi and RoboBus datasets were collected in both urban and suburban environments in Japan.
The RoboTaxi dataset contains about 21,000 frames, and the RoboBus dataset about 13,000 frames.
The annotation schema consists of five classes: Car, Truck, Bus, Bicycle, and Pedestrian.
To the best of our knowledge, this is the first work to construct and experimentally evaluate datasets with different sensor configurations under a unified annotation standard within the same country.

\subsection{Training pipeline}
\label{sec:training}

\textbf{Whole architecture for supervised learning.}
Figure \ref{concept} illustrates the overall architecture for supervised learning.
As the overall training pipeline, we adopt a multi-stage learning strategy based on Downstream Fine-tuning.
For fine-tuning with each sensor configuration, we apply Partial Layer Fine-tuning.
By incorporating these approaches, we obtain models that are optimally tuned for each specific sensor configuration.

\textbf{Whole architecture for unsupervised learning.}
In this study, we also develop an unsupervised domain adaptation pipeline to analyze the domain gap caused by differences in sensor configurations.
Figure \ref{fig:t4dataset} (b) illustrates the overall architecture for unsupervised learning.
As the overall training pipeline, we fine-tune using pseudo-labels generated by an offline model, which uses a more computationally intensive algorithm and larger model parameters to achieve higher performance.

\textbf{Downstream Fine-tuning for Different Sensor Configurations.}
To improve performance for each sensor configuration, we adopt a Downstream Fine-tuning approach as part of a multi-stage training strategy.
First, we perform joint training using all available data across different sensor configurations.
This stage aims to learn generalizable representations and improve robustness across domains.
Subsequently, we apply fine-tuning on each individual sensor configuration dataset to adapt the model to specific domain characteristics.
This step enables the model to specialize in the unique features and distributions of each configuration.

\textbf{Partial Layer Fine-tuning.}
To improve efficiency and performance, we employ Partial Layer Fine-tuning, selectively updating only the layers most effective at bridging domain gaps.
Specifically, we identify and fine-tune layers that are sensitive to differences in sensor configurations, while freezing the remaining layers to reduce computational cost and prevent overfitting.
This approach enables efficient adaptation to domain-specific features.
Recent LiDAR-based 3D object detection models are typically composed of four key modules: encoder, backbone, neck, and head.
In general, the encoder is influenced by input-level properties such as point cloud distribution and intensity patterns.
The head, on the other hand, is affected by output-level characteristics, including object size and annotation formats, as it directly predicts detection targets.
The backbone and neck serve to extract spatial features, and are thus sensitive to viewpoint and sensor placement.
Based on these characteristics, we adopt the following strategy in our setup as detailed in the Appendix.
Since the same mechanical LiDAR is used across configurations and both utilize concatenated point clouds, the encoder experiences similar input distributions and is therefore kept fixed.
In contrast, as sensor layouts differ between configurations, the spatial features extracted by the backbone and neck must adapt accordingly; thus, we make these modules trainable.
Finally, since the annotation protocol is consistent and the nature of the objects to be detected remains the same, the head is also fixed.

\section{Experiment}

%\begin{table}[t]
%\normalsize
%\setlength{\tabcolsep}{4.5 pt}
%\begin{center}
%   \caption{
%      The evaluation summary of RoboTaxi dataset.
%      "RT." means using RoboTaxi dataset.
%      "RB." means using RoboBus dataset.
%      "FT." means our fine-tuning methods.
%      "Tru." means truck.
%      "Bic." means bicycle.
%      "Ped." means pedestrian.
%   }
%\begin{tabular}{|c|c|c|c|c|c|c|c|c|}
%\hline
% RT. & RB. & FT. & mAP & Car & Tru. & Bus & Bic. & Ped. \\
%\hline
%$\checkmark$&     &     & 58.6 & 74.8 & 52.8 & 49.7 & 50.0 & 67.6 \\
%$\checkmark$& $\checkmark$  &     & 63.3 & 74.8 & 53.3 & 66.4 & 56.3 & 65.5 \\
%$\checkmark$& $\checkmark$  & $\checkmark$  & \textbf{64.7} & 75.3 & 50.4 & 70.0 & 62.8 & 65.0 \\
%\hline
%\end{tabular}
%   \label{exp-1}
%\end{center}
%\end{table}
%
%\begin{table}[t]
%\normalsize
%\setlength{\tabcolsep}{4.5pt}
%
%\begin{center}
%   \caption{
%      The evaluation summary of RoboBus dataset.
%      "RT." means using RoboTaxi dataset.
%      "RB." means using RoboBus dataset.
%      "FT." means our fine-tuning methods.
%   }
%\begin{tabular}{|c|c|c|c|c|c|c|c|c|}
%\hline
%RT. & RB. & FT. & mAP & Car & Tru. & Bus & Bic. & Ped. \\
%\hline
%  & $\checkmark$&   & 62.0 & 82.9 & 59.8 & 57.5 & 60.0 & 49.8 \\
%$\checkmark$& $\checkmark$&   & 64.5 & 75.0 & 45.5 & 86.6 & 51.2 & 63.9 \\
%$\checkmark$& $\checkmark$& $\checkmark$& \textbf{65.6} & 75.5 & 46.4 & 89.5 & 54.0 & 62.6 \\
%\hline
%\end{tabular}
%   \label{exp-2}
%\end{center}
%\end{table}

\begin{table*}[t]

\caption{
      The evaluation summary of (a) RoboTaxi dataset and (b) RoboBus dataset.
      "RT." means using RoboTaxi dataset.
      "RB." means using RoboBus dataset.
      "FT." means our fine-tuning methods.
      "Tru." means truck.
      "Bic." means bicycle.
      "Ped." means pedestrian.
}

\normalsize
\setlength{\tabcolsep}{4 pt}

\begin{minipage}[b]{0.48\linewidth}
    \centering
    \begin{tabular}{|c|c|c|c|c|c|c|c|c|}
    \hline
    RT. & RB. & FT. & mAP & Car & Tru. & Bus & Bic. & Ped. \\
    \hline
    $\checkmark$&     &     & 58.6 & 74.8 & 52.8 & 49.7 & 50.0 & 67.6 \\
    $\checkmark$& $\checkmark$  &     & 63.3 & 74.8 & 53.3 & 66.4 & 56.3 & 65.5 \\
    $\checkmark$& $\checkmark$  & $\checkmark$  & \textbf{64.7} & 75.3 & 50.4 & 70.0 & 62.8 & 65.0 \\
    \hline
    \end{tabular}
    \hspace{1.4cm} (a)
\end{minipage}
\begin{minipage}[b]{0.48\linewidth}
    \centering
    \begin{tabular}{|c|c|c|c|c|c|c|c|c|}
    \hline
    RT. & RB. & FT. & mAP & Car & Tru. & Bus & Bic. & Ped. \\
    \hline
    & $\checkmark$&   & 62.0 & 82.9 & 59.8 & 57.5 & 60.0 & 49.8 \\
    $\checkmark$& $\checkmark$&   & 64.5 & 75.0 & 45.5 & 86.6 & 51.2 & 63.9 \\
    $\checkmark$& $\checkmark$& $\checkmark$& \textbf{65.6} & 75.5 & 46.4 & 89.5 & 54.0 & 62.6 \\
    \hline
    \end{tabular}
    \hspace{1.4cm} (b)
\end{minipage}

\label{exp-1}
\end{table*}

%%%

\begin{table*}[t]
\caption{
   Ablation study of Partial Layer Fine-tuning.
   (a) Fine-tuning for RoboTaxi dataset.
   (b) Fine-tuning for RoboBus dataset.
   "En." means encoder.
   "Ba." means backbone.
   "Ne." means neck.
   "He." means head.
}

\normalsize
\setlength{\tabcolsep}{3pt}

\begin{minipage}[b]{0.48\linewidth}
    \centering
    \begin{tabular}{|c|c|c|c|c|c|c|c|c|c|}
    \hline
    En. & Ba. & Ne. & He. & mAP & Car & Tru. & Bus & Bic. & Ped. \\
    \hline
    &   &   &   & 63.3 & 74.8 & 53.3 & 66.4 & 56.3 & 65.5 \\
    &   &   & $\checkmark$& 63.2 & 74.9 & 52.2 & 66.6 & 56.7 & 65.7 \\
    & $\checkmark$&   &   & 64.0 & 75.4 & 50.3 & 69.7 & 59.2 & 65.5 \\
    $\checkmark$& $\checkmark$&   &   & 64.6 & 75.3 & 50.7 & 70.6 & 61.2 & 65.4 \\
    $\checkmark$& $\checkmark$& $\checkmark$&   & 64.3 & 75.3 & 50.7 & 71.6 & 58.2 & 65.6 \\
    & $\checkmark$& $\checkmark$&   & \textbf{64.7} & 75.3 & 50.4 & 70.0 & 62.8 & 65.0 \\
    & $\checkmark$& $\checkmark$& $\checkmark$& 64.2 & 75.0 & 50.0 & 70.8 & 60.0 & 65.1 \\
    $\checkmark$& $\checkmark$& $\checkmark$& $\checkmark$& 64.1 & 75.3 & 50.5 & 70.1 & 58.4 & 65.9 \\
    \hline
    \end{tabular}
    \hspace{0.4cm} (a)
\end{minipage}
\begin{minipage}[b]{0.48\linewidth}
    \centering
    \begin{tabular}{|c|c|c|c|c|c|c|c|c|c|}
    \hline
    En. & Ba. & Ne. & He. & mAP & Car & Tru. & Bus & Bic. & Ped. \\
    \hline
    &   &   &   & 64.5 & 75.0 & 45.5 & 86.6 & 51.2 & 63.9 \\
    &   &   & $\checkmark$& 63.2 & 75.2 & 42.2 & 85.9 & 50.0 & 62.6 \\
    & $\checkmark$&   &   & 65.4 & 75.1 & 46.4 & 90.4 & 52.6 & 62.6 \\
    $\checkmark$& $\checkmark$&   &   & 65.3 & 75.3 & 46.8 & 88.1 & 54.1 & 62.2 \\
    $\checkmark$& $\checkmark$& $\checkmark$&   & 64.8 & 75.7 & 45.7 & 87.9 & 52.6 & 62.3 \\
    & $\checkmark$& $\checkmark$&   & \textbf{65.6} & 75.5 & 46.4 & 89.5 & 54.0 & 62.6 \\
    & $\checkmark$& $\checkmark$& $\checkmark$& 65.2 & 75.1 & 46.5 & 88.0 & 54.7 & 61.9 \\
    $\checkmark$& $\checkmark$& $\checkmark$& $\checkmark$& 65.1 & 75.7 & 45.9 & 87.9 & 54.4 & 61.9 \\
    \hline
    \end{tabular}
    \hspace{0.4cm} (b)
\end{minipage}
\label{exp-3}
\end{table*}

\begin{table*}[t]
\small
\centering
\caption{
   The evaluation results for RoboBus dataset under various training condition and fine-tuning settings.
   "DS." means Downstream Fine-tuning.
   "PL." means Partial Layer Fine-tuning.
   }
\begin{tabular}{|l|c|c|c|c|c|c|c|c|}
\hline
Condition & DS. & PL. & mAP & Car & Tru. & Bus & Bic. & Ped. \\
\hline
RoboTaxi & & & 53.1 & 72.0 & 45.0 & 36.4 & 46.9 & 65.0 \\
RoboTaxi $\rightarrow$ Fine-tuning by RoboBus (10\%) & & & 61.8 & 73.7 & 46.7 & 70.9 & 53.2 & 64.5 \\
RoboTaxi $\rightarrow$ Fine-tuning by RoboBus (20\%) & & & 62.9 & 74.1 & 48.4 & 74.4 & 52.6 & 65.1 \\
RoboTaxi $\rightarrow$ Fine-tuning by RoboBus (50\%) & & & 63.4 & 74.3 & 46.0 & 78.0 & 54.3 & 64.7 \\
RoboTaxi $\rightarrow$ Fine-tuning by RoboBus (100\%) & & & 64.6 & 74.6 & 45.9 & 82.1 & 55.8 & 64.6 \\
\hline
Joint training & & & 64.5 & 75.0 & 45.5 & 86.6 & 51.2 & 63.9 \\
Joint training $\rightarrow$ Fine-tuning by RoboBus & $\checkmark$ & & 65.1 & 75.7 & 45.9 & 87.9 & 54.4 & 61.9 \\
Joint training $\rightarrow$ Fine-tuning by RoboBus & $\checkmark$ & $\checkmark$ & \textbf{65.6} & 75.5 & 46.4 & 89.5 & 54.0 & 62.6 \\
\hline
\end{tabular}
\label{tab:exp-4}
\end{table*}

\begin{table*}[t]
\small
\centering
\caption{
   The evaluation results under unsupervised learning domain adaptation to analyze the domain gap.
   "PL." means Partial Layer Fine-tuning.
   In this experiment, we use pseudo-labels for the RoboBus dataset generated from unlabeled RoboBus data.
   }
\begin{tabular}{|l|c|c|c|c|c|c|c|c|}
\hline
Condition  & PL. & mAP & Car & Tru. & Bus & Bic. & Ped. \\
\hline
RoboTaxi & & 53.1 & 72.0 & 45.0 & 36.4 & 46.9 & 65.0 \\
Joint training & & 64.5 & 75.0 & 45.5 & 86.6 & 51.2 & 63.9 \\
\hline
RoboTaxi $\rightarrow$ Fine-tuning by Pseudo-labeled RoboBus & & 52.0 & 72.0 & 46.1 & 40.5 & 42.6 & 58.6 \\
RoboTaxi $\rightarrow$ Fine-tuning by Pseudo-labeled RoboBus & $\checkmark$ & 52.3 & 72.4 & 46.7 & 39.2 & 44.7 & 58.2\\
\hline
\end{tabular}
\label{tab:exp-5}
\end{table*}

\subsection{Setting}

We used CenterPoint as the 3D object detection model throughout all experiments.
Our CenterPoint consists of encoder of PointPillars \cite{Lang2018PointPillarsFE} as encoder, SECOND \cite{Yan2018SECONDSE} as backbone, FPN of SECOND \cite{Yan2018SECONDSE} as neck, and head of CenterPoint \cite{yin2021center} as head.
Voxel size is \SI{0.32}{\meter} and we do not use multi-frame input (densification).
We conduct experiments on the multi-sensor configuration dataset as described in Section \ref{sec:dataset} and split train, val, and test data for each sensor configuration.
The detection and inference range was set to \SI{120}{\meter}.
For evaluation, we followed the same mAP definition as the nuScenes benchmark.
For the experiments of unsupervised learning, we use the LiDAR-only version of TransFusion \cite{Bai2022TransFusionRL} as the offline model.
Voxel size of the offline model is \SI{0.075}{\meter} and we do not use multi-frame input.

\subsection{Result}

Tables \ref{exp-1} (a) and (b) summarize the evaluation results for supervised learning across different combinations of sensor configurations and the use of fine-tuning.
In Table \ref{exp-1} (a), training with only the RoboTaxi dataset yields a baseline mAP of 58.6.
Incorporating additional data from the RoboBus configuration improves performance to 63.3, demonstrating the benefit of multi-configuration training.
Applying fine-tuning further enhances the mAP to 64.7, with notable gains in the Bus and Bicycle classes, indicating improved cross-domain generalization.
%Similarly, in Table \ref{exp-2}, training with only the RoboBus configuration results in an mAP of 62.0.
Similarly, in Table \ref{exp-1} (b), training with only the RoboBus configuration results in an mAP of 62.0.
When both RoboTaxi and RoboBus datasets are used, the mAP increases to 64.5. With fine-tuning, performance improves further to 65.6, achieving the best results overall.
The most significant improvements are observed in the Bus and Bicycle categories, which are likely more affected by domain shift across sensor setups.
These results confirm that incorporating data from multiple sensor configurations contributes to better generalization, and that fine-tuning effectively adapts the model to diverse domains, leading to consistent performance improvements across classes.
The transition from training on RoboBus only to training on both datasets exhibits a larger performance gap compared to the transition from RoboTaxi only to both datasets.
This discrepancy can be attributed to the greater number of frames in the RoboTaxi dataset, which causes the jointly trained base model to be more heavily influenced by the sensor configuration domain of RoboTaxi.
Our results demonstrate that this RoboTaxi-biased representation can be partially mitigated through downstream fine-tuning, allowing the model to better adapt to the characteristics of other sensor configurations such as RoboBus.

\subsection{Analysis}

Table \ref{exp-3} presents an ablation study of our Partial Layer Fine-tuning method on the RoboTaxi (a) and RoboBus (b) datasets.
We incrementally enable fine-tuning for different modules — encoder, backbone, neck, and head — to assess their individual and combined contributions.
In both settings, fine-tuning only the head slightly degrades performance compared to no fine-tuning, suggesting that isolated tuning of output layers is insufficient for different sensor configuration.
Tuning the backbone leads to improvements, and additional tuning of the encoder and neck further boosts performance.
Interestingly, tuning all modules does not always yield the best result.
The highest mAP on RoboTaxi and RoboBus is achieved when fine-tuning the backbone and neck only, highlighting that selective adaptation of specific layers is more effective than full fine-tuning.
These results support the effectiveness of our partial tuning strategy in balancing adaptation performance with parameter efficiency.
They also demonstrate that naive fine-tuning is not sufficient to achieve optimal performance, whereas Partial Layer Fine-tuning provides a more effective means of adapting to different sensor configurations.

Table \ref{tab:exp-4} presents the evaluation results under various training conditions and fine-tuning settings across different combinations of sensor configurations.
The model trained only on RoboTaxi performs poorly when evaluated on RoboBus, achieving an mAP of just 53.1.
This performance drop is particularly severe in the Bus (36.4) and Truck (45.0) classes, highlighting a substantial domain gap caused by the differences in sensor configurations between RoboTaxi and RoboBus.
To address this issue, we gradually applied fine-tuning using different portions of RoboBus data.
Interestingly, using only 10\% of RoboBus data raised the mAP significantly to 61.8, with the Bus class improving dramatically to 70.9.
As the amount of fine-tuning data increased, performance continued to improve, reaching an mAP of 64.6 when using 100\% of the data, matching the result obtained by joint training.
These results suggest that fine-tuning alone, without the source-domain dataset, can achieve performance comparable to joint training for different sensor configurations.
For the models trained via joint training, we further applied downstream fine-tuning, which improved the mAP from 64.5 to 65.1, surpassing models fine-tuned from the RoboTaxi-only baseline.
Moreover, applying our Partial Layer Fine-tuning method further improved performance to 65.6 mAP.
These findings demonstrate the effectiveness of our fine-tuning strategy, particularly in adapting to sensor-specific domain shifts.

To further investigate the domain gap between different sensor configurations, we conducted an experiment on unsupervised domain adaptation using pseudo-labels generated from the RoboBus dataset without annotations.
Table \ref{tab:exp-5} presents the evaluation results under this setting.
When fine-tuning the RoboTaxi model with pseudo-labeled RoboBus data (RoboTaxi $\rightarrow$ Fine-tuning by Pseudo-labeled RoboBus), the mAP decreased slightly to 52.0.
To mitigate this, we applied Partial Layer Fine-tuning, aiming to avoid overfitting to potentially inaccurate pseudo-labels, and this adjustment led to a minor improvement, raising mAP to 52.3.
Despite this marginal gain, the results remain significantly lower than the RoboTaxi model.
Although the performance of Truck (45.0 → 46.7) and Bus (36.4 → 39.2) slightly improves under the pseudo-label setting, we observe a decline in performance for smaller objects such as Bicycle (46.9 → 44.7) and Pedestrian (65.0 → 58.2).
This suggests that generating accurate pseudo-labels for small objects is particularly challenging, likely due to the limitations of the offline model used for pseudo-labeling.
As noted in \cite{YuanZGYSQC24}, the mAP for three-class domain adaptation from Waymo to nuScenes under an unsupervised setting remains in the teens.
This highlights how unsupervised domain adaptation on public datasets tends to produce low baseline results, framing it more as a detection or discovery task (``can we find the objects?'').
In contrast, the domain gap caused by sensor configuration differences is comparatively narrower.
The focus here shifts beyond simple detection to improving localization and classification accuracy — in other words, from finding to refining.
This distinction supports the motivation behind our fine-tuning strategy, which aims not merely to enable detection in a new domain, but to enhance detection quality by adapting to sensor-specific characteristics.
The results reinforce that our approach to refine for sensor configuration by fine-tuning is well-suited to this refinement phase of domain adaptation.

\section{Conclusion}

In this paper, we addressed domain adaptation in 3D object detection under varying LiDAR sensor configurations.
To address this, we proposed a fine-tuning framework combining Downstream Fine-tuning and Partial Layer Fine-tuning, enabling efficient adaptation to each sensor setup.
We constructed a unified multi-configuration dataset and demonstrated through fine-tuning experiments that our method improves per-configuration performance over joint training.
Ablation studies further highlight that partial fine-tuning offers a better balance between accuracy and efficiency than full model tuning.
Our unsupervised domain adaptation experiments demonstrate that fine-tuning for sensor-specific configurations effectively addresses the narrower domain gap.
We believe this work contributes to making 3D object detectors more adaptable to real-world vehicle platform diversity, helping support broader deployment of autonomous driving systems.

{
    \small
    \bibliographystyle{ieeenat_fullname}
    \bibliography{main}
}

\clearpage
\setcounter{page}{1}
\maketitlesupplementary

\section{Fine-tuning in Large Language Models}

Fine-tuning has traditionally been used in various tasks not only in academia but also in industry.
Recently, with the rapid development of large language models (LLMs) \cite{minaee2025largelanguagemodelssurvey, zhao2025surveylargelanguagemodels}, fine-tuning techniques themselves have also evolved.
After pre-training base models on general natural language data (e.g. GPT \cite{brown2020languagemodelsfewshotlearners, openai2024gpt4technicalreport} or LLaMA \cite{touvron2023llama2openfoundation, grattafiori2024llama3herdmodels}), numerous efforts have been made to adapt these models to specific downstream tasks.
For example, LLaMA-2 has been fine-tuned via supervised fine-tuning (SFT) to create models like LLaMA-2-chat that can follow user prompts; such models are often referred to as instruction-tuned models.
There are also domain-specific fine-tuned models that are optimized for particular fields such as medicine and finance — for instance, Med-PaLM for medical \cite{singhal_large_2023}, or FinGPT \cite{yang2023fingptopensourcefinanciallarge} for finance.
These developments have brought renewed attention to the use of fine-tuning as a method for specialization.
This study is inspired by such fine-tuning tasks in the LLM domain.

\section{Additional analysis for fine-tuning}
\subsection{The Total Number of Parameters}

\begin{figure}[t]
   \begin{center}
      \includegraphics[width=0.99\linewidth]{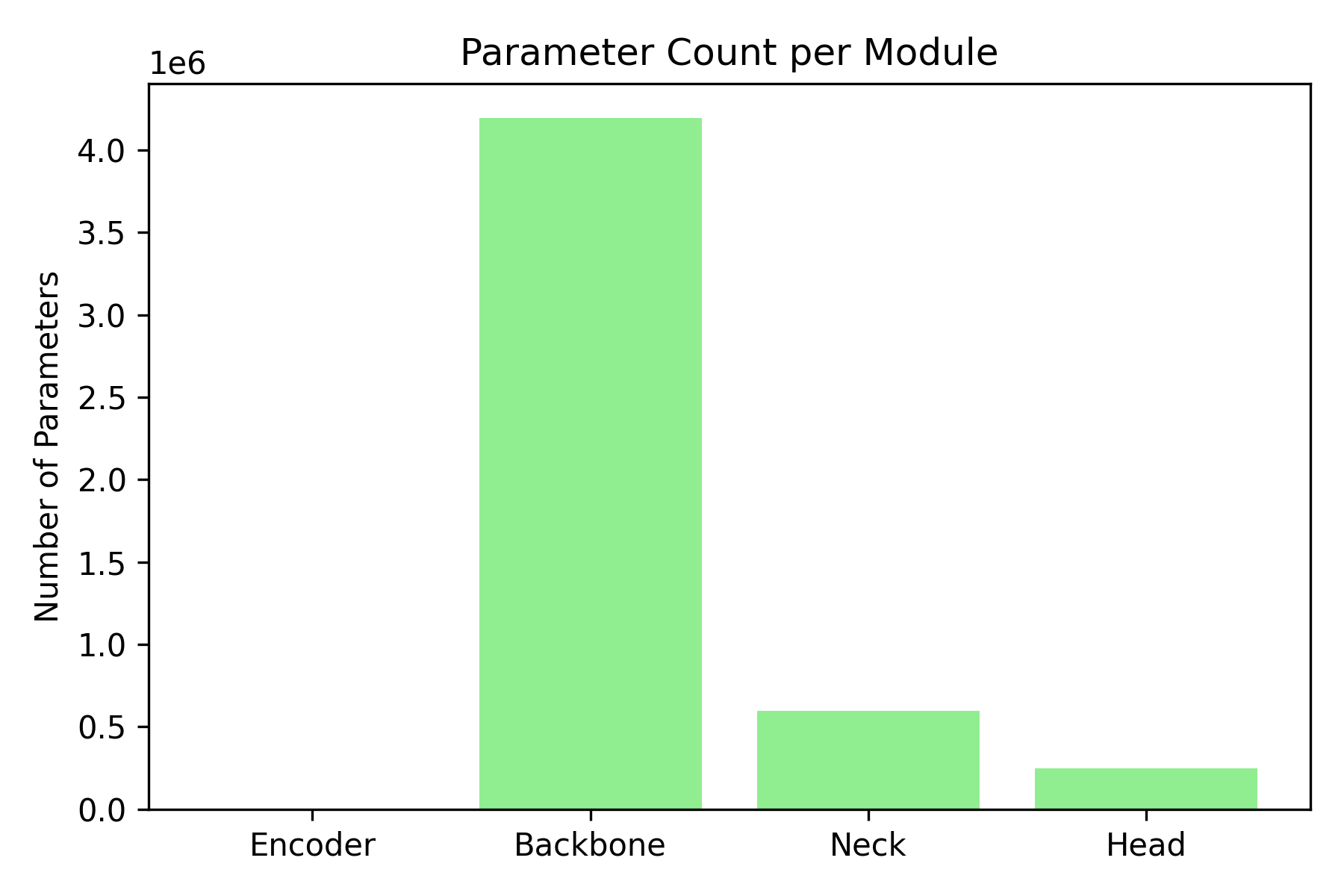}
   \end{center}
   \caption{
    The total number of parameters for each layer.
   }
   \label{fig:param}
\end{figure}

We first examined the total number of parameters in the model.
The parameter count for each architectural module is as follows:

\begin{itemize}
\item Encoder: 1,360 parameters
\item Backbone: 4,194,304 parameters
\item Neck: 599,552 parameters
\item Head: 248,527 parameters
\end{itemize}

We illustrate this in Figure \ref{fig:param}.
This breakdown provides a structural understanding of where the model's capacity is concentrated and serves as a basis for interpreting the magnitude of parameter differences across modules.

\subsection{Analysis methods}

\begin{algorithm}[t]
\caption{Compute Layer-wise Parameter Difference}
\SetKwInOut{Input}{Input}
\SetKwInOut{Output}{Output}
\SetKwComment{Comment}{$\triangleright$\ }{}

\Input{Base model parameters $\mathcal{B}$, Target model parameters $\mathcal{T}$, \\
\hspace{2.3em} Mode flag $\text{is\_relative}$ (boolean)}
\Output{Layer-wise L1 difference $\mathcal{D}$ for each module}

Initialize $\mathcal{D}_\text{diff}[\text{Encoder, Backbone, Neck, Head}] \gets 0.0$ \\
Initialize $\mathcal{D}_\text{sum}[\text{Encoder, Backbone, Neck, Head}] \gets 0.0$ \\

\ForEach{parameter name $k \in \mathcal{B}$}{
    \If{$k \in \mathcal{T}$ \textbf{and} $\mathcal{B}[k]$ is floating point}{
        $l \gets$ \texttt{CategorizeLayer}$(k)$ \\
        $\delta \gets \lVert \mathcal{B}[k] - \mathcal{T}[k] \rVert_1$ \\
        $s \gets \lVert \mathcal{B}[k] \rVert_1$ \\
        $\mathcal{D}_\text{diff}[l] \gets \mathcal{D}_\text{diff}[l] + \delta$ \\
        $\mathcal{D}_\text{sum}[l] \gets \mathcal{D}_\text{sum}[l] + s$ \\
    }
}

\ForEach{layer type $l$ in \{\text{Encoder, Backbone, Neck, Head}\}}{
    \If{$\text{is\_relative} = \text{True}$ \textbf{and} $\mathcal{D}_\text{sum}[l] > 0$}{
        $\mathcal{D}[l] \gets \mathcal{D}_\text{diff}[l] / \mathcal{D}_\text{sum}[l]$ \Comment*[r]{Relative difference}
    }
    \Else{
        $\mathcal{D}[l] \gets \mathcal{D}_\text{diff}[l]$ \Comment*[r]{Absolute difference}
    }
}

\Return{$\mathcal{D}$}
\label{layer_diff}
\end{algorithm}

To analyze parameter-level changes across model components, we compute the L1 difference between two sets of model weights.
Algorithm \ref{layer_diff} illustrates the procedure for calculating the layer-wise parameter differences.
The algorithm first categorizes each parameter into one of four architectural modules: Encoder, Backbone, Neck, or Head, based on the parameter name.
For each matched parameter between the two models, we compute the L1 distance and accumulate both the absolute difference and the L1 norm of the original parameter values.
The final output can be computed in two modes.

\begin{itemize}
\item Relative difference: the accumulated L1 difference is normalized by the L1 norm of the original parameters, yielding a scale-invariant measure.
\item Absolute difference: the accumulated raw L1 difference is returned, which reflects the total magnitude of change regardless of scale.
\end{itemize}

The computation mode is controlled via a boolean flag is\_relative. When set to True, the algorithm returns relative differences; otherwise, absolute values are used. This flexibility allows for analyzing both raw parameter shifts and their proportional impacts, depending on the application context.

\subsection{Analysis Results}

\begin{figure}[tb]
   \begin{center}
      \includegraphics[width=0.99\linewidth]{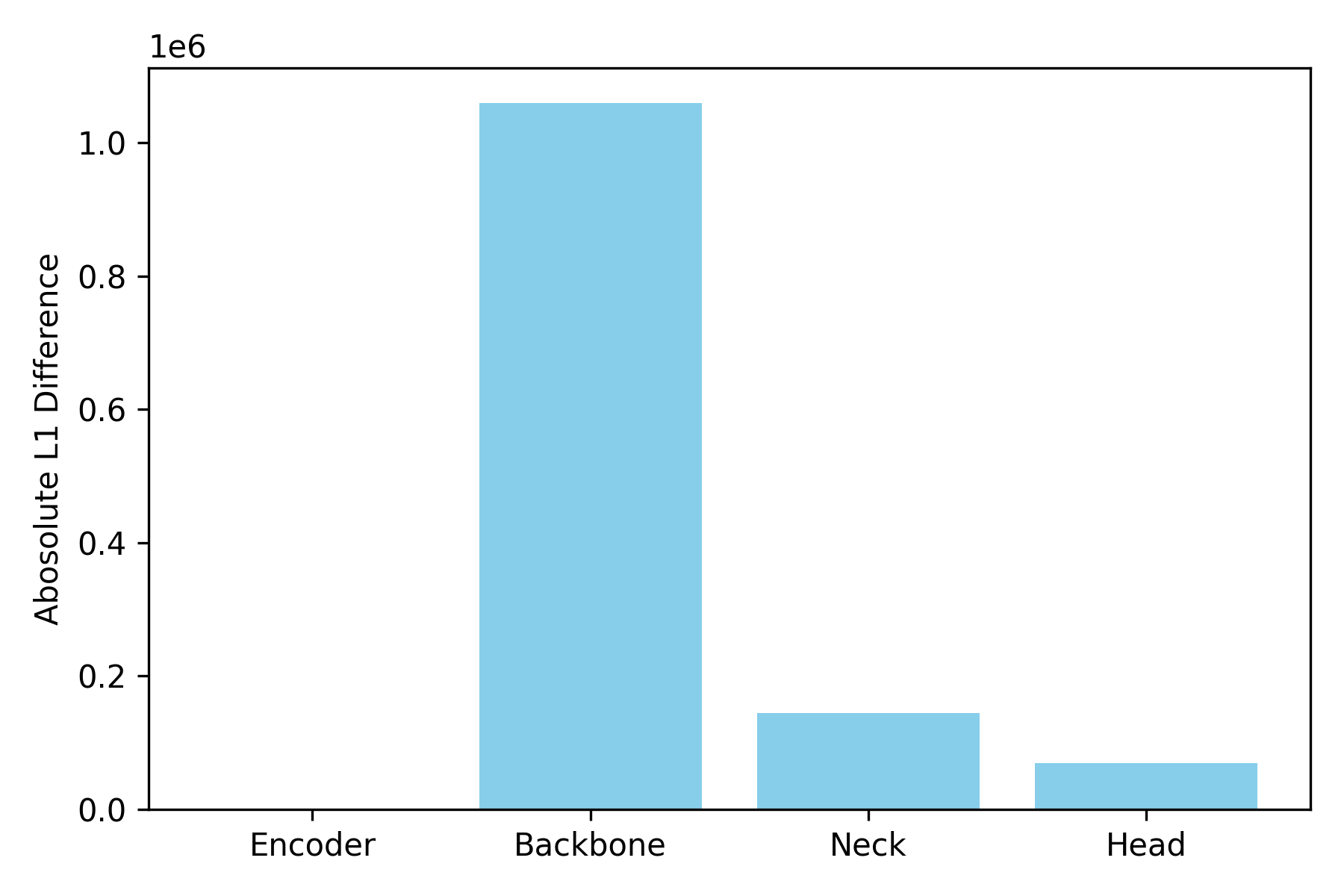}
   \end{center}
   \caption{
     The absolute parameter differences between the RoboTaxi model and the RoboBus model.
   }
   \label{analysis-1}
\end{figure}

\begin{figure}[tb]
   \begin{center}
      \includegraphics[width=0.99\linewidth]{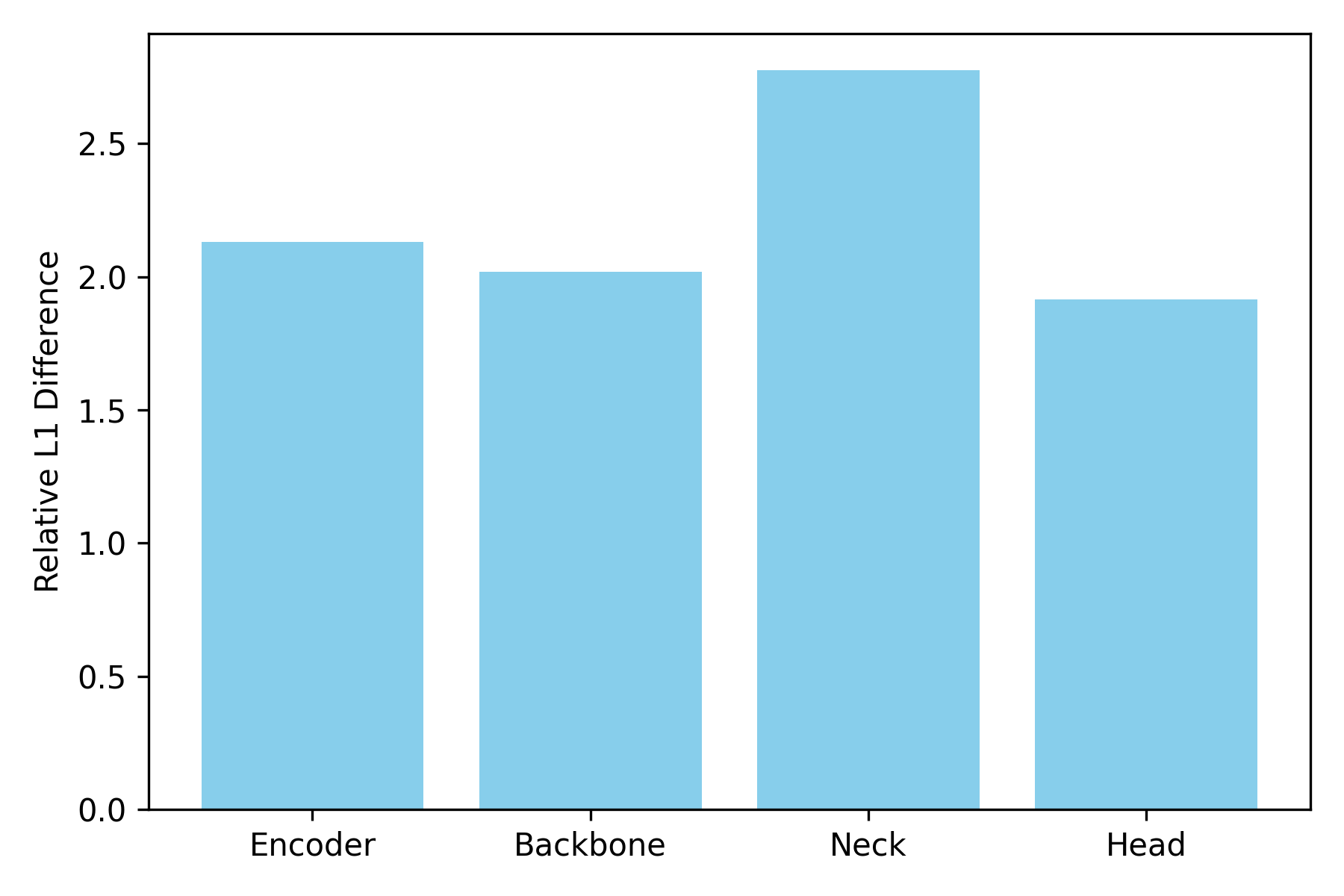}
   \end{center}
   \caption{
      The relative parameter differences between the RoboTaxi model and the RoboBus model.
   }
   \label{analysis-2}
\end{figure}

\begin{figure}[tb]
   \begin{center}
      \includegraphics[width=0.99\linewidth]{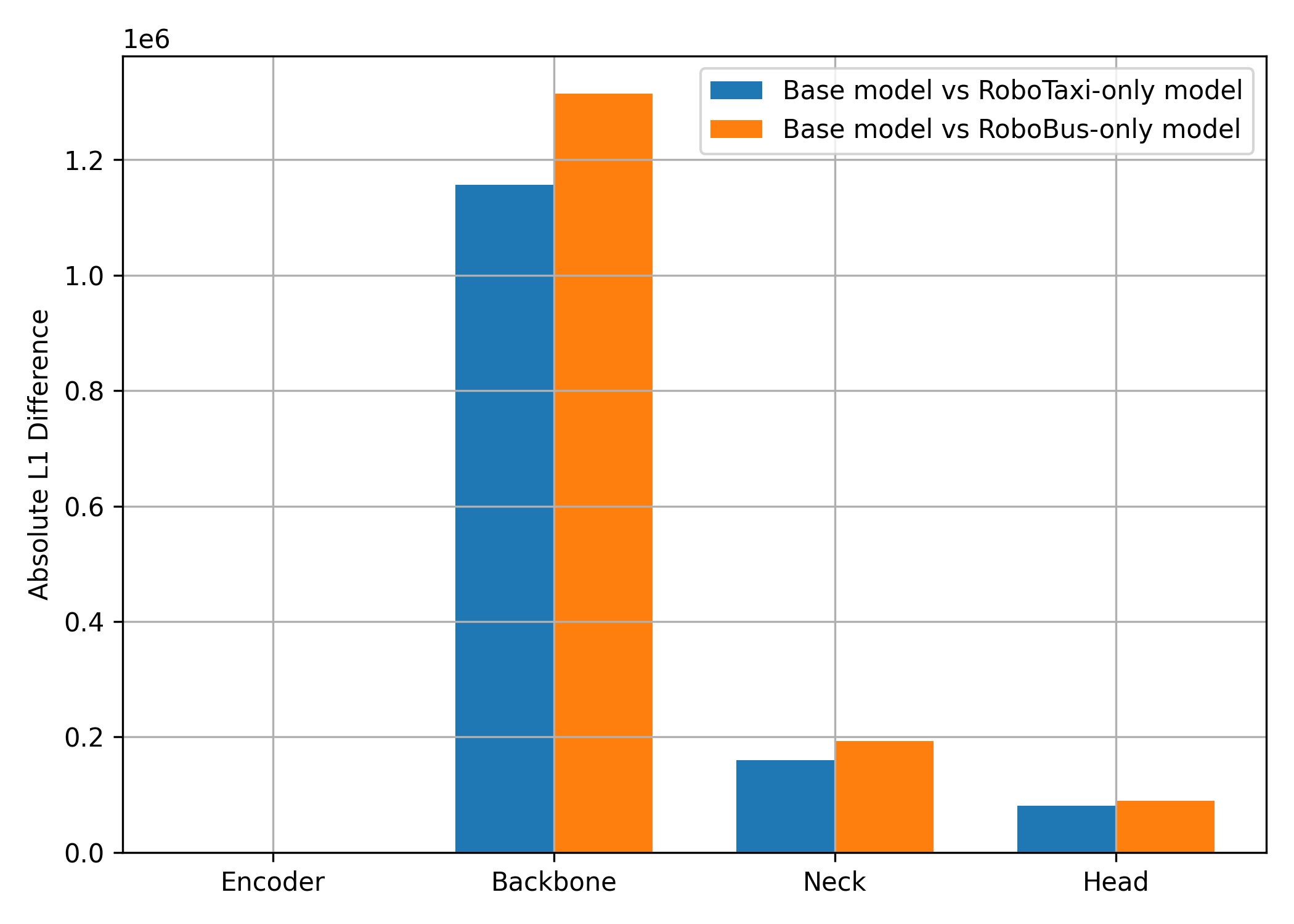}
   \end{center}
   \caption{
      The absolute differences between the base model and RoboTaxi model, and between the base model and RoboBus model.
   }
   \label{analysis-3}
\end{figure}

\begin{figure}[tb]
   \begin{center}
      \includegraphics[width=0.99\linewidth]{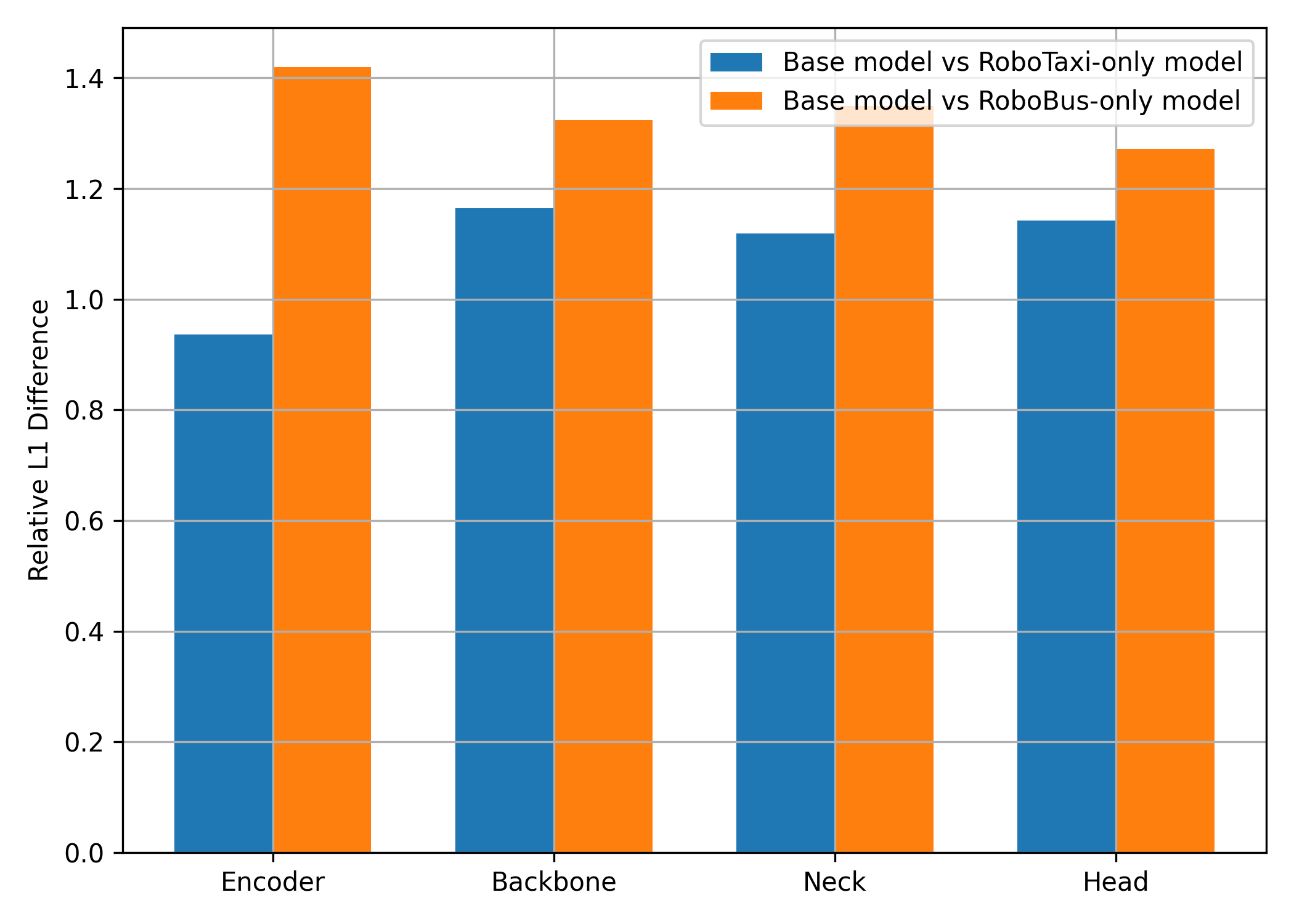}
   \end{center}
   \caption{
      The relative differences between the base model and RoboTaxi model, and between the base model and RoboBus model.
   }
   \label{analysis-4}
\end{figure}

To better understand how different modules are affected by domain shift and training regimes, we analyzed the parameter differences across models.
We first examined the absolute parameter differences between the RoboTaxi model and the RoboBus model.
As shown in Figure \ref{analysis-1}, the differences across modules are largely proportional to the number of parameters each contains.
This indicates that, in terms of raw magnitude, modules with more parameters naturally exhibit larger absolute shifts during training.

To remove the effect of parameter count and focus on proportional changes, we next analyzed the relative parameter differences between the same models, as shown in Figure \ref{analysis-2}.
Interestingly, the Neck module shows the largest relative deviation, suggesting that it is most affected by the domain gap stemming from different sensor configurations.
Since the Neck is responsible for aggregating multi-scale features from the backbone, this result implies that feature representation in this stage is particularly sensitive to sensor-induced variations.

We further examined the differences between the base model — obtained through joint training on both RoboTaxi and RoboBus datasets — and the domain-specific models.
Figure \ref{analysis-3} illustrates the absolute parameter differences between the base model and each of the domain-specific models.
The RoboTaxi model is noticeably closer to the base model in absolute terms, which can be attributed to the fact that the RoboTaxi dataset contributes a greater number of frames to the joint training process.
As a result, the base model is more strongly influenced by RoboTaxi data during optimization.

Relative differences between the base model and each of the domain-specific models are presented in Figure \ref{analysis-4}.
These results further confirm that the RoboTaxi model remains closer to the base model, not only in absolute terms but also when normalized by parameter magnitude.
The Encoder module shows the largest relative deviation, which can be explained by its small parameter count — minor changes have a disproportionately large effect on the relative metric.
Conversely, the Head module shows the smallest relative change, suggesting that object categories and prediction structures are shared across domains.
This implies that the domain gap caused by sensor configuration primarily affects early to mid-level representations, while the output space remains relatively stable.

\end{document}